\documentclass[journal]{IEEEtran}
\usepackage{graphicx}
\usepackage{amsmath}
\usepackage{booktabs}
\usepackage{array}
\usepackage{xcolor}
\usepackage{amssymb}
\usepackage{multicol}
\usepackage{float}
\usepackage{stfloats}
\usepackage{array}
\usepackage{cite}    
\makeatletter
\let\NAT@parse\undefined
\makeatother
\usepackage[colorlinks,linkcolor=blue,citecolor=blue]{hyperref}

\begin{document}
\title{Contrastive Learning-Driven Traffic Sign Perception: Multi-Modal Fusion of Text and Vision}
\author{Qiang~Lu, Waikit~Xiu, Xiying~Li~\IEEEmembership{Member, IEEE}, Shenyu~Hu, Shengbo~Sun 

\thanks{WaiKit~Xiu and Qiang~Lu contributed equally to this work. 

This work is going to be submitted to the IEEE for possible publication. Copyright may be transferred without notice, after which this version may no longer be accessible.

This work was supported by the National Natural Science Foundation of China (Grant No. U21B2090); the 2024 Higher Education Scientific Research Project of Guangzhou Municipal Education Bureau (Grant No. 2024312014). (Corresponding author: Xiying Li.)

Qiang Lu, Waikit Xiu, Xiying Li, Shenyu Hu and Shengbo Sun are with the School of Intelligent Systems Engineering, Sun Yat-sen University, Shenzhen 518107, China, and also with Guangdong Provincial Key Laboratory of Intelligent Transportation System, Guangzhou 510275, China (e-mail: xiuwk0820@connect.hku.hk;luqiang8@mail2.sysu.edu.cn and stslxy@mail.sysu.edu.cn).
}
}

\maketitle

\begin{abstract}
Traffic sign recognition, as a core component of autonomous driving perception systems, directly influences vehicle environmental awareness and driving safety. Current technologies face two significant challenges: first, the traffic sign dataset exhibits a pronounced long-tail distribution, resulting in a substantial decline in recognition performance of traditional convolutional networks when processing low-frequency and out-of-distribution classes; second, traffic signs in real-world scenarios are predominantly small targets with significant scale variations, making it difficult to extract multi-scale features.To overcome these issues, we propose a novel two-stage framework combining open-vocabulary detection and cross-modal learning. For traffic sign detection, our NanoVerse YOLO model integrates a reparameterizable vision-language path aggregation network (RepVL-PAN) and an SPD-Conv module to specifically enhance feature extraction for small, multi-scale targets. For traffic sign classification, we designed a Traffic Sign Recognition Multimodal Contrastive Learning model (TSR-MCL). By contrasting visual features from a Vision Transformer with semantic features from a rule-based BERT, TSR-MCL learns robust, frequency-independent representations, effectively mitigating class confusion caused by data imbalance.
On the TT100K dataset, our method achieves a state-of-the-art 78.4\% mAP in the long-tail detection task for all-class recognition. The model also obtains 91.8\% accuracy and 88.9\% recall, significantly outperforming mainstream algorithms and demonstrating superior accuracy and generalization in complex, open-world scenarios.
\end{abstract}


\IEEEpeerreviewmaketitle

\section{Introduction}
 \IEEEPARstart{A}{corrding} to the World Health Organization (WHO), road traffic accidents cause about 1.35 million deaths globally each year, and more than 90\% of these accidents are related to human driving errors. Autonomous driving technology is regarded as an important solution to reduce the rate of traffic accidents, and its safety has become the core bottleneck that restricts the development of the technology\cite{WHO_report}. Traffic Sign Recognition (TSR) mainly relies on the visual perception module of the automatic driving system [some articles can be cited here], and provides the core basis for vehicle behavioral decision-making and safe driving by accurately parsing the key information such as road instructions, warnings, and prohibitions. Therefore, realizing high-precision and real-time traffic sign recognition is not only the cornerstone for guaranteeing the safety and reliability of the automatic driving system, but also of great practical significance for enhancing the overall effectiveness of the intelligent transportation system.

With the advancement of artificial intelligence, Internet of Things technologies, and edge computing, significant strides have been made in the field of intelligent driving \cite{intro1,intro2,intro3}. A number of studies [please supplement specific references] have laid a crucial foundation for visual technical solutions in traffic sign recognition. However, vision-based traffic sign recognition tasks still confront numerous challenges. Traffic sign recognition is specifically divided into two stages: detection and classification. In the detection stage, traffic signs, characterized by their small size and blurred features in images, may lead to missed detections, thereby compromising the safety of autonomous driving systems. Recent studies have demonstrated that although many deep learning-based detection algorithms, such as the YOLO series and Faster R-CNN, achieve promising performance on standard datasets, they still suffer from a significant drop in detection rates when dealing with small targets\cite{intro_small1,intro_small2}. Moreover, these convolution-based algorithms are often confined to the limited scenarios within the training sets, resulting in frequent missed detections in actual open traffic scenes\cite{intro6,intro7}.In the classification stage, current research on traffic sign recognition generally assumes that the class distributions of the training and test sets are consistent and that samples are sufficient. Many studies\cite{Huige} only retain categories with more than 100 instances in the dataset for training and evaluation. Despite some studies\cite{intro_4,intro_5} optimizing for the long-tail problem, such efforts are mostly limited to single adjustments of algorithm parameters or loss mechanisms. They fail to overcome the inherent flaw of single modalities, which merely mechanically memorize image features and lack a deep understanding of the semantic connotations of traffic signs. Consequently, when confronted with novel or variant signs, they become ineffective due to their inability to parse the underlying semantic logic.

To address the aforementioned challenges in traffic sign recognition, this paper decouples the task into two phases: detection and classification, proposing a traffic sign recognition algorithm that integrates open vocabulary detection with cross-modal contrastive learning. In the detection phase, an open vocabulary detection algorithm, NanoVerse-YOLO, is introduced to achieve feature extraction of small signs and accurate localization of traffic signs in open scenarios. In the classification phase, a cross-modal contrastive learning framework is designed, combining visual encoders and text encoders to compute the cosine similarity of the dual-modal features, enabling fine-grained semantic matching of open categories and calculating classification results through probability distribution functions. However, existing research lacks data for aligning the semantics of traffic sign images and texts, making effective training of this framework difficult. To address this, we have independently constructed a dedicated dataset that includes semantic annotations for open category texts, completing the pre-training of the cross-modal contrastive learning framework. By deeply integrating spatial localization information from the detection phase with semantic understanding capabilities from the classification phase, we build a traffic sign recognition system with a collaborative reasoning mechanism, significantly improving recognition accuracy and model generalization in complex open scenarios. Therefore, the contributions of this paper can be summarized as follows:

\begin{itemize}
    \item  A two-stage traffic sign recognition framework based on open vocabulary detection and cross-modal contrastive learning is constructed, significantly enhancing the model's semantic understanding capability of traffic signs by integrating visual images and textual semantic information.
    
    \item  A multi-modal collaborative annotation framework and dynamic prompt template are designed for the traffic sign recognition task, providing a standardized semantic-visual alignment benchmark for open vocabulary understanding.
    
    \item The Rule-BERT encoder is proposed to structurally parse traffic sign texts and retain key semantic tuples, providing effective semantic representation for traffic sign understanding.
\end{itemize}

\section{Related Work}
\subsection{Traffic Sign Recognition}
In recent years, with the rise of deep learning, convolutional neural network (CNN)-based object detection methods have achieved significant progress in the field of traffic sign detection\cite{Re3,Re4}. Two-stage detectors, such as Faster R-CNN\cite{FasterRCNN} and its variants, as well as one-stage detectors like the YOLO\cite{Re6} series and SSD\cite{Re7}, are widely applied in traffic sign detection tasks. For example, Li X et al.\cite{Re6}, Yu J et al. \cite{Re7}, and You S et al. \cite{FasterRCNN}, have proposed improved algorithms for traffic sign detection based on Faster R-CNN, YOLO, and SSD, respectively. However, the inherent small size of traffic signs and the severe long-tail distribution of class samples in real traffic scenarios remain two core challenges facing current traffic sign detection, significantly limiting the robustness and generalization capability of models.

According to common standards in the field of object detection, the determination of small targets is primarily based on two criteria: the COCO dataset categorizes detection targets smaller than 32×32 pixels as small targets, while SPIE \cite{SPIE} defines small targets as those whose area occupies less than 0.12\% of the total image area. A statistical analysis of the representative traffic sign dataset TT100K reveals that among 26,361 labeled bounding boxes, the number of samples meeting the small target criterion (<32×32 pixels) reaches 9,973, accounting for 37.8\% of the total samples. Small target detection is a significant challenge in traffic sign recognition, prompting researchers to propose various technical solutions. Some researchers have adopted multi-scale feature fusion strategies. For instance, Tang Q et al. (2020) \cite{Re7} proposed an improved FPN-based network that enhances the detection performance of small traffic signs by integrating finer-grained features; Liu Y et al. (2021) \cite{FasterRCNN} designed a context-aware feature fusion module that utilizes semantic information from surrounding areas to enhance the representation of small signs. However, for extremely small traffic signs, even multi-scale fusion struggles to fully compensate for the loss of original information. Inspired by generative networks, Dewi C et al. (2021) \cite{Re9} proposed a feature reconstruction network aimed at recovering details of small targets in deep feature maps, while Chen L C et al. (2017) \cite{Re10} employed dilated convolutions to expand the receptive field and capture contextual information of small targets. Data augmentation is a common means to alleviate the deficiency and lack of diversity in small target samples. Horn D et al. (2022) \cite{Re11} proposed a "copy-paste" augmentation strategy for traffic signs, which involves copying small traffic signs to different locations to increase the number of instances. Additionally, some studies have optimized the training process by adjusting anchor box sizes \cite{Re12} or designing loss functions specifically for small targets \cite{Re13}. Introducing attention mechanisms can also prompt networks to focus more on small target regions and their key features. Liu F et al. (2020) \cite{Re14} proposed a spatial-channel attention module to adaptively enhance the feature response of small traffic signs, while Min K et al. (2023)\cite{Re15} explored how to refine features of different scales through adaptive weight allocation in the feature pyramid to meet the detection needs of small targets.

There exists a severe long-tail problem in the sample distribution of various categories of traffic signs. For example, signs such as speed limits and no entry appear frequently, while special warning signs like wildlife crossing and icy road conditions are extremely rare. Taking the TT100K dataset as an example, its construction accurately reflects real traffic environments, but the number of instances for different categories shows extreme skewness, with some categories appearing only in the training set while others appear only in the test set, leading to cross-set shifts. This makes it difficult for traditional CNN algorithms to effectively capture the discriminative features of sparse categories, resulting in significant performance degradation of classifiers in long-tail distribution scenarios. To address the long-tail problem, research has primarily focused on the following areas: Re-sampling is a direct approach. Buda M et al. (2020) \cite{Re16} proposed a general method for balancing training data by oversampling minority classes and undersampling majority classes, while Wang Y et al. (2021) \cite{Re17} introduced an adaptive resampling strategy that dynamically adjusts sampling rates based on class counts and learning progress. However, oversampling may lead to model overfitting on noise from minority samples, while undersampling may result in the loss of important information. Data generation is also an effective means, with some methods using Generative Adversarial Networks (GANs) or diffusion models to synthesize high-quality images of rare traffic signs \cite{Re18,Re19} to augment training data. Additionally, loss function re-weighting balances the optimization process by adjusting the weights of different classes in the loss function. The class-balanced loss proposed by Cui et al. (2019) \cite{Re20} has been successfully applied to traffic sign detection tasks \cite{Re21}, while the loss function introduced by Lin et al. (2020) \cite{Re22} combines the advantages of Focal Loss by incorporating class-sensitive adjustment factors. Finally, decoupled learning is also viewed as an effective strategy. Kang et al. (2019) \cite{Re24} proposed a framework that first trains a feature extractor and then uses a classifier designed specifically for long-tail classification. In traffic sign detection, Tan J et al. (2023) \cite{Re25} explored methods to separate feature learning and classification head training within the detection framework to enhance the recognition of rare traffic signs. The combination of these methods can more effectively address the challenges posed by long-tail distributions.

The above review indicates that significant progress has been made in the field of traffic sign detection in terms of small target recognition and long-tail problem handling. In the aspect of small target detection, relevant feature extraction enhancement strategies and loss function improvements provide important references for the method proposed in this paper. Regarding the long-tail problem, current solutions such as resampling and data generation, which rely on single-modal data, struggle to fully capture the effective features of rare categories due to their insufficient samples and vulnerability to interference from complex environments. Moreover, it is even more challenging to accurately mine and match features for unknown categories. Therefore, this paper adopts a cross-modal learning approach, integrating multi-modal information by constructing a high-dimensional semantic space to more effectively infer the features of categories with few instances, thereby significantly improving the recognition ability for rare categories.

\subsection{Cross-Modal Learning Method}
In recent years, multimodal learning has made significant breakthroughs in the integration of computer vision and natural language processing, particularly demonstrating strong generalization capabilities in tasks such as image understanding and visual question answering. The CLIP\cite{CLIP} model, for example, employs an innovative dual-modal contrastive training mechanism that maps visual information and language descriptions into a unified feature space. By utilizing large-scale image-text pair data, it brings semantically similar images and texts closer together in the embedding space while distancing heterogeneous content, thereby constructing a universal framework for cross-modal understanding. Inspired by this, visual language models like OpenAI's GPT-4o\cite{GPT4}, Google's Gemini\cite{Gemini}, and the DeepSeek\cite{DeepSeekV3} series have emerged, excelling in general image-text understanding and generation tasks. Additionally, multimodal technology has been successfully applied in fine-grained downstream tasks such as medical diagnosis and industrial inspection. For instance, Li et al\cite{ZeroShotECG}. combined electrocardiograms with medical records for disease classification, further validating its effectiveness in specific fields and providing new technical insights for Traffic Sign Recognition (TSR) tasks.

In the field of traffic sign recognition, some researchers have explored cross-modal learning methods to enhance model recognition capabilities for rare categories. For example, Cao et al\cite{Cao2022}. proposed a TSR method based on zero-shot learning that effectively identifies out-of-training-set sign categories by integrating visual features with semantic descriptions. Gan et al.\cite{Gan2023} utilized the GPT-4o model to extract semantic information and combined it with visual features to complete classification tasks, preliminarily validating the application potential of multimodal methods in this context. However, these methods still exhibit significant limitations in practical applications. Firstly, they are highly dependent on high-end computing devices or large model APIs, leading to excessive inference response delays that cannot meet the stringent real-time decision-making requirements of autonomous driving scenarios. Secondly, existing general-purpose large models are often designed for general tasks and lack specialized optimization and fine-tuning for fine-grained tasks like traffic sign recognition. This makes it challenging to adapt to the characteristics of small traffic sign targets, high similarity, and complex environments, resulting in poor accuracy in recognizing rare or novel signs and failing to meet the high-precision perception demands in autonomous driving scenarios.

This study aims to address the shortcomings of current work by proposing a unified and collaborative framework to efficiently tackle the issues of small target missed detection and long-tail distribution in traffic sign detection. We introduce NanoVerse-YOLO, a deep multi-scale feature enhancement algorithm for open-vocabulary detection, specifically designed to capture the subtle features of small-sized traffic signs and locate signs in open scenes. By proposing a fine-grained classification model based on cross-modal contrastive learning, both aspects collectively improve the robustness of traffic sign detection and the recognition accuracy of rare small categories, providing a more reliable approach for intelligent recognition of traffic signs in real-world traffic scenarios.

\section{Methodology}
\begin{figure*}[htbp]
    \centering
    \includegraphics[width=\textwidth,trim=14cm 5.5cm 7cm 5cm, clip]{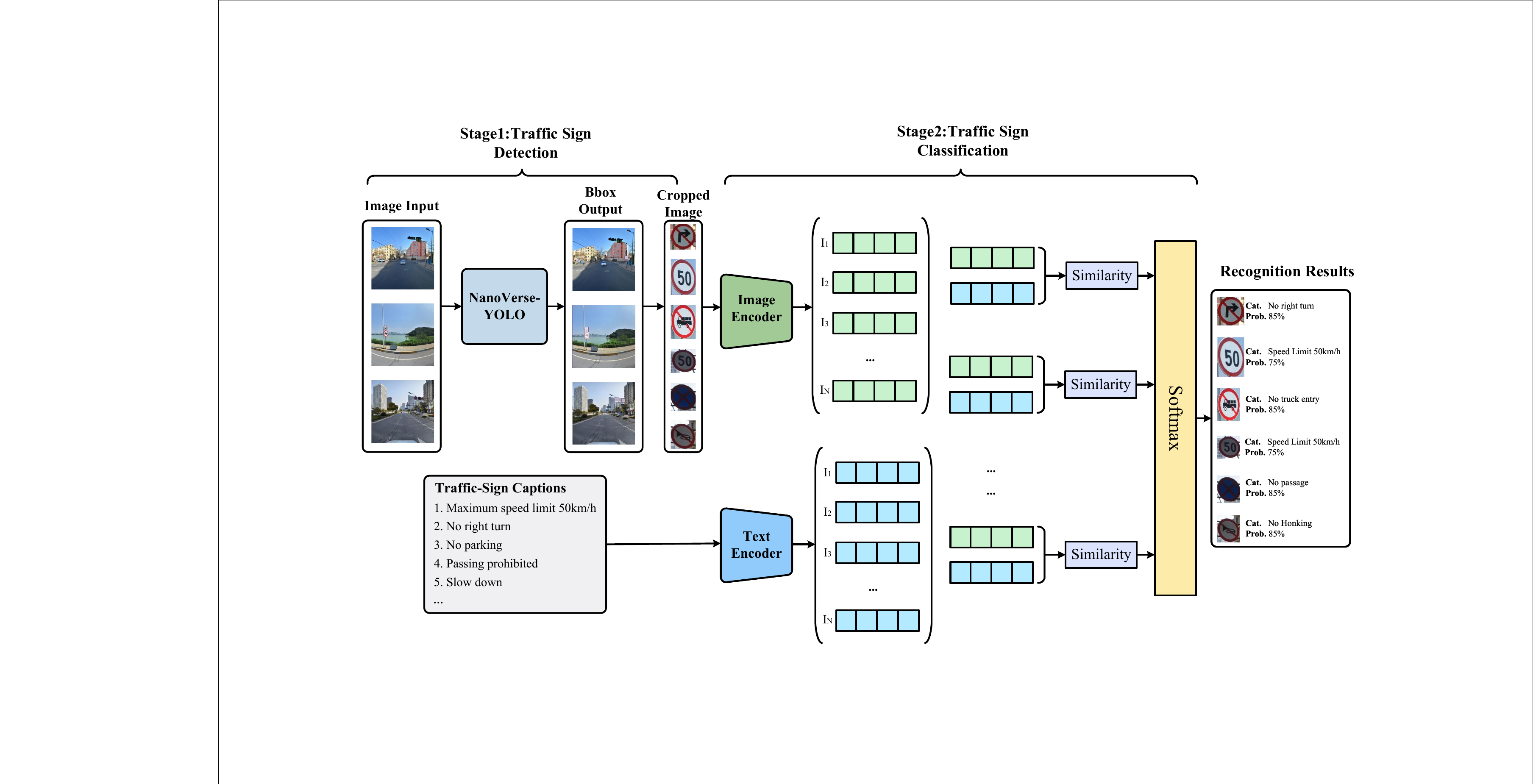} 
    \caption{Coarse-to-Fine Traffic Sign Recognition Framework. The framework is divided into two stages: Stage 1 involves Traffic Sign Detection using the NanoVerse-YOLO algorithm to process input images, output bounding boxes, and crop individual sign images; Stage 2 involves Traffic Sign Classification, where image features and text semantic vectors are extracted using an image encoder and text encoder, respectively. The recognition results are generated based on calculated similarities.}
    \label{framework}
\end{figure*}
This paper presents a Coarse-to-Fine Recognition Framework for traffic sign recognition. This framework is constructed using an open vocabulary detection algorithm enhanced for small targets and a classifier based on cross-modal contrastive learning. The algorithmic flow is illustrated in Figure 1. In the target detection phase, the OpenYOLO-NanoVerse algorithm processes the input images to output bounding boxes for traffic sign locations, followed by cropping individual sign images. In the subsequent classification phase, image features and text semantic vectors are extracted using a visual encoder and the Rule-BERT text encoder, respectively. After similarity calculations and Softmax processing, the recognition results are outputted.Furthermore, the research is based on the TT100K dataset and a self-constructed Traffic Sign Text-Image Alignment Dataset (TSTIAD). The TSTIAD consists of 24,715 individual sign images with fine-grained text labels, covering 221 categories, and is divided into training and testing sets in a 2:1 ratio. This dataset is standardized to meet the requirements of cross-modal learning models, providing high-quality data support for visual-text alignment within the framework.

\subsection{NanoVerse-YOLO: Traffic Sign Open Vocabulary Detector}
\begin{figure*}[htbp]
    \centering
    \includegraphics[width=0.8\textwidth, trim=1.4cm 14.5cm 7.6cm 8cm, clip]{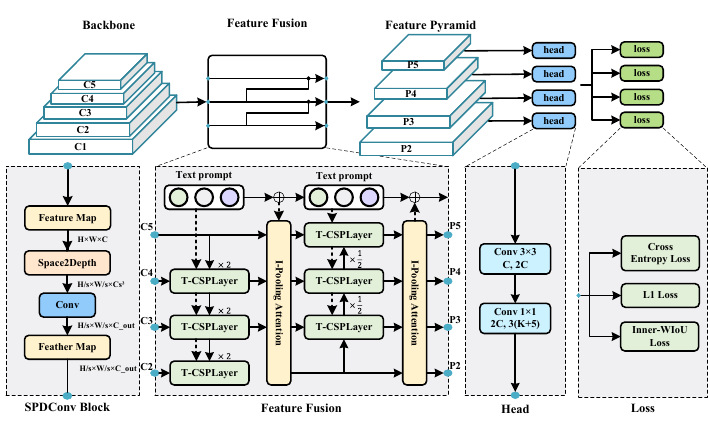}
    \caption{This diagram shows the architecture of the NanoVerse-YOLO model, including the Backbone for feature extraction, Feature Fusion using T-CSPLayer, and the Loss components that optimize the model with various loss functions.}
    \label{Detector}
\end{figure*}
To tackle the challenge of small target detection for traffic signs and achieve precise localization of targets in open scenarios, this paper proposes the OpenYOLO-NanoVerse algorithm. The algorithm integrates a re-parameterizable visual-language path aggregation network (RepVL-PAN), deeply interacting visual and textual features to enhance the detection capability of open scene targets. It designs the SPD-Conv module to replace traditional Conv modules, balancing feature map resolution and detail retention; introduces the P2 small detection head to enhance small target response; and adopts the Inner-WIoU loss function to accurately evaluate target position and overlap quality, improving small target localization performance.

RepVL-PAN consists of two key modules. The T-CSPLayer module injects textual features into certain layers across stages, achieving multi-modal information fusion from top to bottom or bottom to top. For multi-scale image features \(X\) and text embeddings \(W\), the textual guidance signal is integrated into the image features via \(X'_i = \text{Concat}\left(X_i, \sigma\left(X_i, W^T\right)\right)\), where the activation function \(\sigma(\cdot)\) dynamically adjusts the scale feature weights, introducing flexible text-guided information on the multi-scale feature maps. The I-Pooling Attention module pools image features into \(3 \times 3\) regions at each scale, utilizing a multi-head attention mechanism for feature interaction. In the calculation formula \(W' = W + \text{MultiHeadAttention}(W, x, X)\), the downsampled image features \(X\) and text features \(W\) interact to compute the updated text embedding \(W'\), capturing global context and enhancing the visual perception of text embeddings. During the inference phase of the algorithm, offline vocabulary embeddings are re-parameterized as convolutional or linear layer weights, simplifying model deployment.

To address the difficulties in small target detection for traffic signs, this paper specifically designs visual feature extraction. In the backbone network, traditional Conv modules cause significant reductions in the spatial resolution of feature maps due to strided convolutions and pooling operations, leading to information loss, which can be represented as 
\begin{equation}
\mathcal{L}_{info} = \sum_{i=1}^{s^2} \mathbb{I}(\mathbf{X}_{sub}^{(i)} \notin \mathbf{Y}_{down}) \cdot \|\mathbf{X}_{sub}^{(i)}\|_2
\end{equation}
where \(s\) is the downsampling stride, \(\mathbf{X}_{sub}\) is the sub-region of the input feature map, and \(\mathbf{Y}_{down}\) is the downsampled output, which is unfavorable for small target detection. Therefore, the SPD-Conv module is employed, consisting of an SPD layer and a non-strided convolution layer. For the input feature map \(\mathbf{X} \in \mathbb{R}^{H \times W \times C}\), the SPD layer first rearranges it spatially into a high-dimensional representation \(\mathbf{Y} = \text{reshape}(\mathbf{X}, \frac{H}{s}, s, \frac{W}{s}, s, C)\), where \(\mathbf{Y} \in \mathbb{R}^{\frac{H}{s} \times \frac{W}{s} \times (C \cdot s^2)}\), and then outputs features through the non-strided convolution layer as \(\mathbf{Y_{conv}} = \phi(\mathbf{W}^T \mathbf{Y} + \mathbf{b})\), where \(\phi\) is the LeakyReLU activation function. Meanwhile, a P2 small detection head is added at the model's head to introduce multi-scale information, enhancing the response capability for small targets.

In the loss function design, the Inner-WIoU is adopted, expressed as 
\begin{equation}
L_\text{Inner-WIoU} = L_\text{WIoU} + \text{IoU} - \text{IoU}^\text{inner}
\end{equation}
integrating \(L_\text{WIoU}\) with \(\text{IoU}^\text{inner}\). Here, \(\text{IoU}^\text{inner} = \frac{\text{inter}}{\text{union}}\) is used to evaluate the quality of the overlapping area between the predicted and ground truth boxes; 
the Wise-IoU expressed as
\begin{equation}
L_\text{WIoU} = \gamma \left(\frac{(p_x - g_x)^2}{w_g^2} + \frac{(p_y - g_y)^2}{h_g^2}\right)
\end{equation}
it enhances the model's sensitivity to target location by considering the positional deviations between the predicted and ground truth boxes. By combining these two loss functions, the Inner-WIoU loss function effectively enhances the model's performance in small target detection.

\subsection{Traffic Sign Text-Image Alignment Dataset}
\begin{figure*}[htbp]
    \centering
    \includegraphics[width=\textwidth, trim=11cm 15.5cm 15cm 3cm, clip]{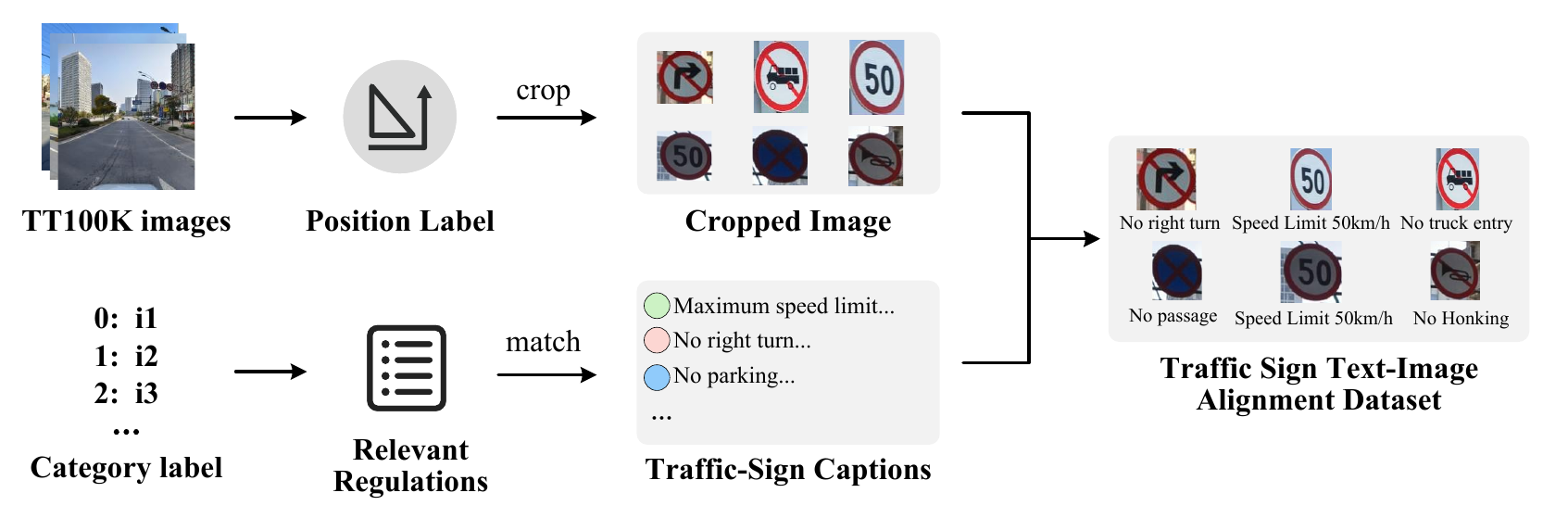} 
    \caption{This diagram illustrates the process of aligning traffic sign images with their corresponding textual descriptions, including cropping the images based on position labels and matching them with relevant regulations.}
    \label{data}
\end{figure*}
Traffic signs serve as efficient visual symbols that map to traffic regulations, ensuring intuitive and interpretable information transmission. The shape, color, patterns, and other visual elements of each sign are closely related to their corresponding traffic rules, reflecting the inherent alignment between normative text and visual information. This makes cross-modal contrastive learning an ideal approach for traffic sign research. However, existing datasets such as TT100K and the German traffic sign dataset label visual localization in a simplistic manner using letters and numbers, completely overlooking the deep connections between visual and semantic aspects. They lack effective characterization of descriptive attributes and sub-category features of the signs, making them insufficient to meet the demands of cross-modal learning tasks.

To fill this research gap and provide high-quality data support for the model training proposed in Chapter C, we introduce a text description expansion method aimed at generating refined, high-quality descriptive attributes for each traffic sign category. Based on the TT100K dataset, we expand the text labels for cross-modal alignment tasks, constructing a new dataset that includes over 24,715 images covering 221 categories, equipped with fine-grained traffic rule text descriptions. The specific production process is as follows:

\subsubsection{Text Description Generation} Based on the Chinese Road Traffic Regulations (GB5768-2022), we crafted precise and detailed text descriptions for each traffic sign category. For example, category A1 is described as "a circular blue sign with a white arrow indicating straight ahead." Such detailed descriptions not only enhance the accuracy of the signs but also improve their usability in machine learning.

\subsubsection{Image Cropping and Pairing} Using the original annotation boxes, we cropped the traffic sign areas to generate 24,715 single-sign images, matching them with their corresponding text descriptions. This step ensures the high quality and consistency of the dataset, providing a solid foundation for subsequent model training.

\subsubsection{Dataset Division and Formatting} The final dataset is divided into training and testing samples in a 2:1 ratio, comprising 16,477 pairs for training and 8,238 pairs for testing, while strictly maintaining the original category distribution. The dataset underwent standardization to align with mainstream cross-modal learning models' training requirements.

This dataset achieves precise alignment between traffic sign visual features and regulatory text for the first time, providing an important foundation for cross-modal learning research. By offering a rich correspondence between visual and textual information, we enhance the model's understanding of traffic signs, providing new insights and directions for traffic sign recognition research.

\subsection{Multimodal Contrastive Learning for Fine-Grained Classification}
To achieve fine-grained classification of traffic signs, this paper proposes a multi-modal contrastive learning framework. The framework employs a dual-encoder architecture, using Vision Transformer (ViT) as the visual encoder and Rule-BERT as the text encoder. It implements deep feature alignment between visual and textual modalities through contrastive learning methods.

In terms of visual feature extraction, the ViT architecture explicitly models the global semantic relationships of images through a self-attention mechanism. Its unique image tokenization method efficiently transforms the input image \( I \in \mathbb{R}^{H \times W \times C} \) into a sequence representation similar to text tokens. Specifically, the image is divided into fixed-size patches, flattened, and mapped into vectors, yielding a patch embedding sequence \( \{p_1, p_2, \ldots, p_N\} \) (where \( N = \frac{H \times W}{p^2} \), and \( p \) is the length of the patch). Positional encoding is added to retain spatial information. The embedding sequence is then fed into a multi-layer Transformer encoder, where each layer captures global features using multi-head self-attention \( \text{Attention}(\cdot) \) and multi-layer perceptron \( \text{MLP}(\cdot) \). The output is updated in the \( l \)-th layer according to 
\begin{equation}
Z^{(l+1)} = \text{LayerNorm}(\text{Attention}(Z^{(l)}) + Z^{(l)})
\end{equation}
and
\begin{equation}
Z^{(l+1)} = \text{LayerNorm}(\text{MLP}(Z^{(l+1)}) + Z^{(l+1)})
\end{equation}

Finally, ViT uses a special \( [CLS] \) token to extract the global image feature \( f_v \in \mathbb{R}^d \), enabling precise matching with text features in a shared embedding space for subsequent cross-modal alignment tasks.

The design of the text encoder, Rule-BERT, addresses the shortcomings of traditional BPE tokenizers when handling traffic sign texts. Traditional tokenizers often misinterpret key information (e.g., splitting speed limit numbers) and fail to utilize the structured and repetitive characteristics of texts. To mitigate this, we introduce rule-based regular expressions and a traffic regulation knowledge base into the BPE tokenizer. Regular expressions parse the text and extract key semantic tuples, combined with a number protection mechanism to prevent key information, such as speed limits, from being split. Specifically, the input text \( T \) is first segmented into sub-word level token sequences \( X = [x_1, x_2, \dots, x_n] \) and mapped to embedding representations \( E = [e_1, e_2, \dots, e_n], e_i \in \mathbb{R}^d \). This is then input into a multi-layer Transformer encoder. In each layer of the Transformer, the query, key, and value matrices are computed as $Q^{(l)} = E^{(l)} W_Q, \quad K^{(l)} = E^{(l)} W_K, \quad V^{(l)} = E^{(l)} W_V$, and processed through the self-attention mechanism 
\begin{equation}
\text{Attention}(Q, K, V) = \text{softmax}\left(\frac{QK^T}{\sqrt{d}}\right)V
\end{equation}

With residual connections and layer normalization, the output is updated as 
\begin{equation}
E^{(l+1)} = \text{LayerNorm}(\text{Attention}(Q^{(l)}, K^{(l)}, V^{(l)}) + E^{(l)})
\end{equation}

Ultimately, the special token \( [CLS] \) extracts the global semantic representation of the text sequence \( f_t \in \mathbb{R}^d \). During inference, to enhance efficiency, the results are stored in a semantic cache space after the first encoding, avoiding redundant computations.

After the visual encoder and text encoder complete the feature extraction for images and texts, cross-modal feature alignment and classification decisions become crucial for improving recognition performance. For the input image-text pair \( \{v_i, t_i\} \), the initial features \( f'_{v,i} \) and \( f'_{t,i} \) are obtained through the visual and text encoders. These are mapped to a shared embedding space via \( L_2 \) normalization. A similarity matrix is constructed as 

\begin{equation}
S_{i,j} = \hat{f}_{v,i} \cdot (\hat{f}_{t,j})^\top
\end{equation}

to measure semantic relevance. To further optimize cross-modal alignment, a learnable temperature scaling factor \( \tau = \exp(\gamma) \) is introduced, leading to a bi-directional contrastive loss function: 

\begin{equation}
\begin{split}
L_{\text{cross-modal}} = -\frac{1}{2B} \Bigg( \sum_{i=1}^{B} \log \frac{\exp(\tau S_{i,i})}{\sum_{j=1}^{B} \exp(\tau S_{i,j})} + \\
 \sum_{i=1}^{B} \log \frac{\exp(\tau S_{i,i})}{\sum_{j=1}^{B} \exp(\tau S_{j,i})} \Bigg)
\end{split}
\end{equation}

This function minimizes the loss, thereby reducing the distance between matched pairs while increasing the distance between unmatched pairs, achieving semantic decoupling between modalities. In the classification decision phase, based on the optimized similarity matrix \( S \), the probability distribution 

\begin{equation}
 P(t_j | v_i) = \frac{\exp(\tau S_{i,j})}{\sum_{k=1}^{B} \exp(\tau S_{i,k})}
\end{equation}

is calculated using the \( Softmax \) function. This transforms the cross-modal alignment results into class predictions and provides theoretical support for the inference and validation of traffic sign classification.
\begin{figure*}[htbp]
    \centering
    \includegraphics[width=\textwidth, trim=17cm 6cm 15cm 14cm, clip]{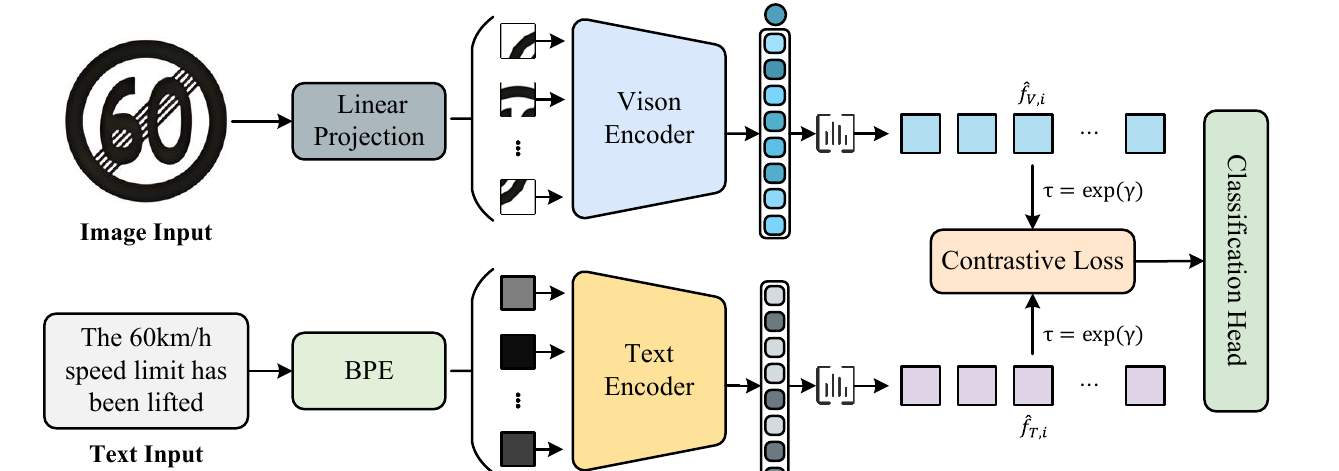} 
    \caption{This diagram illustrates the multi-modal contrastive learning framework, depicting the image and text input processing through respective encoders, followed by a linear projection and the calculation of contrastive loss to optimize the classification head.}
    \label{CLIP}
\end{figure*}

\section{Experiment}
\subsection{Dataset and Evaluation Metrics}
\subsubsection{Dataset} This study is based on the publicly available TT100K dataset, jointly developed by Tsinghua University and Tencent, which contains over 100,000 images, with more than 10,000 images meticulously annotated, covering information on sign positions, categories, and shapes. Building on this, we introduce a self-constructed Traffic Sign Text-Image Alignment Dataset, adding fine-grained traffic regulation text labels to the data. The training set of the dataset includes 144 annotated categories, exhibiting a significant long-tail distribution: 31 head categories (with instance counts > 100), 43 middle categories (10-100 instances), and 69 tail categories (with instance counts < 10). There are also extreme distribution scenarios where A categories appear only in the training set and B categories appear only in the test set. During model training, the first-stage detection module uses only images and location labels, while the second-stage classification module employs only images and text labels.
\subsubsection{Valuation Metrics}
This paper adopts core metrics from the field of object detection to evaluate the performance of traffic sign recognition models, including Precision, Recall, and Mean Average Precision (mAP). 
Precision measures the reliability of detection results, calculated as:
\begin{equation}
    P = \frac{TP}{TP + FP}
\end{equation}
Recall evaluates the ability to cover the target, defined as:
\begin{equation}
R = \frac{TP}{TP + FN}
\end{equation}
where \(TP\) is the number of true positives (predicted boxes with IoU $\geq$ 0.5 and correct classification), \(FP\) is the number of false positives (predicted boxes with IoU < 0.5 or duplicate detections), and \(FN\) is the number of missed true signs.
Mean Average Precision is calculated using two criteria: the single-threshold \(mAP@50\) is computed as:
\begin{equation}
AP_{50}^{(c)} = \int_{0}^{1} p_c(r) \, dr \approx \frac{1}{11}\sum_{k=0}^{10} \max_{\tilde{r}\geq \frac{k}{10}} p_c(\tilde{r})
\end{equation}
The multi-threshold \(mAP@50:95\) calculates \(AP_{\tau}(c)\) for \(\tau \in \{0.5, 0.55, \ldots, 0.95\}\) and takes the average. The final mAP is the comprehensive value of the average precision for each category, given by:
\begin{equation}
mAP = \frac{1}{C}\sum_{c=1}^{C} \left[ \frac{1}{10}\sum_{\tau} AP_{\tau}^{(c)} \right]
\end{equation}

\subsection{Experimental Setup and Details}
The experiment was conducted on a workstation equipped with an NVIDIA RTX 3090 graphics card, using the PyTorch framework. The input images for the first-stage detection model were set to 640×640 pixels, with a batch size of 64. The training process employed the Adam optimizer, with an initial learning rate set to 3e-4, and was trained for a total of 200 epochs. For the second-stage recognition model, the input images were uniformly adjusted to a resolution of 224×224 pixels, with a batch size of 32. The training process also used the Adam optimizer with the same initial learning rate of 3e-4 and was trained for 200 epochs. To enhance training efficiency, mixed precision training techniques were enabled.

\subsection{Experiment Result on TT100K}
To evaluate the performance of the model in the traffic sign detection task, four representative algorithms from the YOLO series were selected: YOLOv8s, YOLOv11s, YOLOv12s, and YOLO-TS. The comparison experiments focused on four core metrics: Precision, Recall, mAP50 (mean Average Precision at an IoU threshold of 0.5), and mAP50:95 (mean Average Precision averaged over the IoU threshold range of 0.5 to 0.95). These experiments were conducted on the TT100K traffic sign dataset to quantitatively assess the differences in model performance regarding detection accuracy and adaptability to complex scenarios. The experimental results are presented below:
\begin{table}[h]
    \centering
    \caption{Comparison of Traffic Sign Detection Models}
    \begin{tabular}{@{}lcccccc@{}}
        \toprule
        Algorithm & Year & Precision & Recall & mAP50(\%) & mAP50:95 \\ 
        \midrule
        YOLOv8s & 2023 & 70.6\% & 57.8\% & 64.4\% & 51.9\% \\
        YOLO11s & 2024 & 74.2\% & 55.6\% & 65.3\% & 52.3\% \\
        YOLO12s & 2024 & 64.0\% & 61.1\% & 65.1\% & 52.1\% \\
        YOLO\_TS & 2025 & 68.8\% & 61.5\% & 66.4\% & 55.9\% \\ 
        \midrule
        Ours & 2025 & 91.8\% & 88.9\% & 78.4\% & 71.9\% \\ 
        \bottomrule
    \end{tabular}
    \label{tab:traffic_sign_detection}
\end{table}

The experimental results fully validate that the model proposed in this paper significantly outperforms mainstream algorithms (such as YOLOv8s, YOLOv11s, etc.) on core metrics for traffic sign detection, demonstrating a notable performance advantage. The Precision reached 91.8\%, significantly leading other models; the Recall was 88.9\%, highlighting the model's exceptional ability in localizing small targets and accurately classifying different categories. The mAP50 reached 78.4\%, and the mAP50:95 reached 71.9\%, showcasing the model's strong control over detection performance across different confidence thresholds. This effectively demonstrates the validity and superiority of the proposed method in the traffic sign detection task.
\begin{figure}[htbp]
    \centering
    \includegraphics[width=0.48\textwidth]{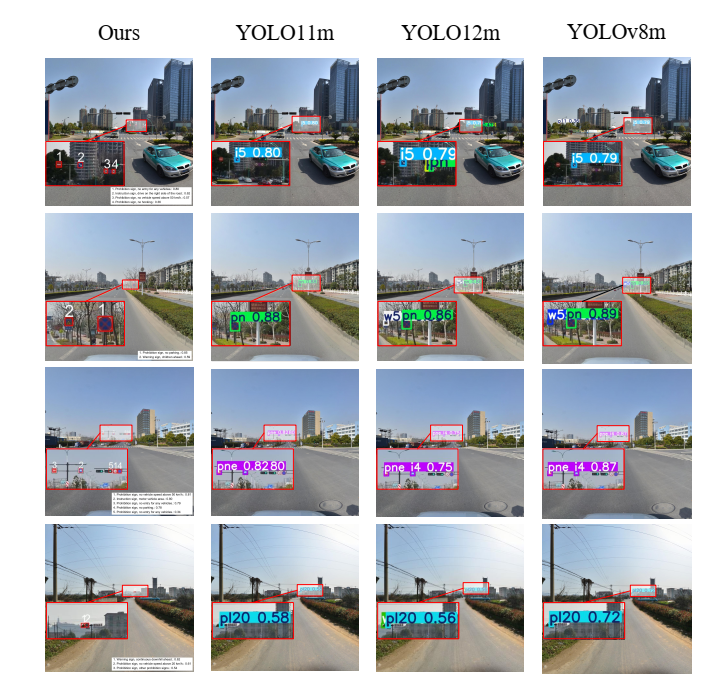} 
    \caption{Visualization of Comparison Experiment Results}
    \label{compare}
\end{figure}
From the comparison results above, it is clear that the algorithm proposed in this paper significantly outperforms other mainstream algorithms such as YOLO11m, YOLO12m, and YOLOv8m in small object detection and fine-grained classification. Furthermore, our algorithm accurately conveys the traffic rule information represented by the traffic signs. In complex scenarios (such as distant or partially obscured traffic signs), our algorithm can accurately capture small targets and demonstrates higher precision and robustness in classification. In contrast, other comparative algorithms often experience missed detections or false positives for small objects, and their performance in classifying long-tail categories is relatively weak. This validates the effectiveness of the specially designed small object detection head and the optimized feature extraction module within our model. Therefore, our algorithm exhibits stronger adaptability and detection accuracy in complex traffic scenarios, effectively addressing the long-tail problem to some extent.

\begin{table*}[htbp]
    \centering
    \setlength{\tabcolsep}{4mm}{
    \caption{Comparison of Our Method with State-of-the-Art Methods on Long-Tail Categories in the TT100K Dataset}
    \footnotesize 
    \renewcommand{\arraystretch}{1.2} 
    \begin{tabular}{p{3.75cm}p{1cm}p{1cm}p{1cm}p{1cm}p{1cm}p{1cm}p{1cm}}
        \toprule
        \textbf{Methods} & \textbf{i1} & \textbf{i4} & \textbf{p2} & \textbf{p3} & \textbf{p4} & \textbf{p35} & \textbf{p70} \\
        \midrule
        YOLOv5\cite{YOLOv5}          & \textbf{8.3} & 8.3  & 3.8 & 33.2 & \textbf{99.5} & 1.5  & 5.0 \\
        Faster R-CNN\cite{FasterRCNN}   & 0.0 & 0.0  & 0.0 & 0.0 &  0.0 & 0.0   & 76.8 \\
        Cascade R-CNN\cite{Cascade} & 0.0 & 40.0 & 0.0 & 0.0 &  0.0 & 0.0   & 0.0 \\
        ATSS\cite{ATSS}            & 0.0 & 40.0 & 16.8 & 0.0 & 1.6 & 0.8   & 0.0 \\
        ZENG et al\cite{Zeng}      & 2.6 & 16.6 & 50.2 & \textbf{34.2} & \textbf{99.5} & 4.3 & 13.0 \\
        \midrule
        \textbf{Ours}              & 2.9 & \textbf{92.7} & \textbf{71.0} & 0.0 & 0.0 & \textbf{100.0} & \textbf{100.0} \\
        \bottomrule
    \end{tabular}}
    \label{tab:comparison}
\end{table*}

\begin{table*}[htbp]
    \centering
    \setlength{\tabcolsep}{4mm}{
    \footnotesize
    \renewcommand{\arraystretch}{1.2} 
    \begin{tabular}{p{3.75cm}p{1cm}p{1cm}p{1cm}p{1cm}p{1cm}p{1cm}p{1cm}}
        \toprule
        \textbf{Methods} & \textbf{pr80} & \textbf{pw3} & \textbf{w35} & \textbf{il70} & \textbf{w12} & \textbf{pw4} & \textbf{pw3.5} \\
        \midrule
        YOLOv5\cite{YOLOv5}          & 0.7  & 49.8 & 99.5 & 42.4 &  0.0 & 50.9  & 62.2 \\
        Faster R-CNN\cite{FasterRCNN}  & 0.0  & 0.0  & 0.0 &  42.5 &  0.0 & 0.0   & 0.0 \\
        Cascade R-CNN\cite{Cascade} & 0.0  & 0.0  & \textbf{100.0} & 40.9 & 0.0 & 0.0   & 9.8 \\
        ATSS\cite{ATSS}            & 48.2 & 1.9  & 90.0 & 34.6 &  0.0 & 20.3  & 45.9 \\
        ZENG et al\cite{Zeng}     & 15.3 & 99.5 & 99.5 & 73.7 &  2.3 & 66.5  & 79.7 \\
        \midrule
        \textbf{Ours}              & \textbf{100.0} & \textbf{100.0} & 25.0 & \textbf{100.0} & \textbf{96.7} & \textbf{100.0} & \textbf{99.5} \\
        \bottomrule
    \end{tabular}}
    \label{tab:comparison1}
\end{table*}

\begin{table*}[htbp]
    \centering
    \setlength{\tabcolsep}{4mm}{
    \footnotesize
    \renewcommand{\arraystretch}{1.2} 
    \begin{tabular}{p{3.75cm}p{1cm}p{1cm}p{1cm}p{1cm}p{1cm}p{1cm}p{1cm}}
        \toprule
        \textbf{Methods} & \textbf{w16} & \textbf{w20} & \textbf{i1} & \textbf{ph2.2} & \textbf{pm15} & \textbf{w45} & \textbf{p8} \\
        \midrule
        YOLOv5\cite{YOLOv5}          & 6.3 & 72.7  & \textbf{49.8} & 58.8 & 14.5 & 76.5  & 33.9 \\
        Faster R-CNN\cite{FasterRCNN}   & 4.8 & 23.6  & 0.0  & 19.0 & 1.2 & 26.5   & 4.1 \\
        Cascade R-CNN\cite{Cascade} & 4.3 & 26.9 & 0.0  & 23.6 & 9.5 & 23.2   & 5.1 \\
        ATSS\cite{ATSS}            & 5.6 & 13.5 & 0.0  & 9.8 &  5.4 & 25.4  & 36.5 \\
        ZENG et al\cite{Zeng}      & 68.7 & \textbf{99.5} & \textbf{49.8} & 61.9 & 13.2 & 75.1 & \textbf{42.0} \\
        \midrule
        \textbf{Ours}              & \textbf{100.0} & 33.3 & 0.0 & \textbf{100.0} & \textbf{70.5} & \textbf{95.0} & 11.1 \\
        \bottomrule
    \end{tabular}}
    \label{tab:comparison2}
\end{table*}

\subsection{Long-Tail Category Detection Performance}
To further validate the model's performance on long-tail categories, this paper conducts specialized experiments on rare categories in the TT100K dataset with fewer than 10 instances. During the experiments, the mAP@0.5 metric for each algorithm is precisely calculated for each category. A comprehensive comparison is made between our method and classic object detection algorithms such as YOLOv5, Faster R-CNN, Cascade R-CNN, and ATSS, as well as algorithms designed for long-tail classification, such as those by ZENG et al. The experimental results are presented below.

The experimental results show that classic object detection algorithms such as YOLOv5 and Faster R-CNN have significant limitations in long-tail category detection tasks. For rare categories in the TT100K dataset with fewer than 10 instances, both Faster R-CNN and Cascade R-CNN exhibit low mAP@0.5 values in multiple categories, such as i1 and p2. This indicates that traditional algorithms struggle to capture the features of low-frequency categories and have inherent flaws when dealing with long-tail distribution issues. In comparison to algorithms specifically designed for long-tail problems, our method demonstrates advantages in most categories. In the pr80 category, our method achieves an mAP@0.5 value of 100.0, far exceeding the 15.3 of similar algorithms. In the i4 category, our method leads significantly with an mAP@0.5 value of 92.7. For the p2 and p3 categories, our method achieves scores of 100.0 and 71.0, respectively, showing a marked advantage over similar algorithms with scores of 50.2 and 34.2. In other categories like pw3, our method also consistently maintains an mAP@0.5 value of 100.0. Although in some categories, like w35, our method has an mAP@0.5 value of 25.0, which is slightly lower than some similar algorithms, in the vast majority of categories within the long-tail detection task of the TT100K dataset, our method's mAP@0.5 significantly outperforms other algorithms. This fully demonstrates that our method possesses higher detection accuracy and stronger recognition capability in addressing the long-tail category detection problem for traffic signs, making it a more effective solution for this issue.

\subsection{Ablation Study}
To verify the effectiveness of cross-modal learning in the traffic sign recognition task, this paper conducts a progressive ablation study based on the NanoVerse-YOLO localization model. The experiment adopts a strategy that progresses from simple to complex, using a single-stage detector as the baseline and gradually adding components such as a classification module trained with fine-grained text labels, a rule-enhanced BPE tokenizer, and a semantic cache. This study systematically explores the impact of each component on the model's performance based on the TT100K dataset. The experimental results are presented below.
\begin{table*}[htbp]
    \centering
    \caption{Ablation Study Results}
    \renewcommand{\arraystretch}{1.5}
    \begin{tabular}{@{}lcccccccc@{}}
        \toprule
        \textbf{Model} & \textbf{Classifier} & \textbf{Rule-BERT} & \textbf{Feature Cache} & \textbf{P} & \textbf{mAP50} & \textbf{mAP50:95} & \textbf{FPS} \\
        \midrule
        NanoVerse-YOLO &  &  &  & 75.8\% & 27.0\% & 30.3\% & 2.36 \\
        NanoVerse-YOLO & \checkmark &  &  & 87.4\% & 84.6\% & 73.6\% & 65.7 \\
        NanoVerse-YOLO & \checkmark & \checkmark &  & 91.8\% & 88.9\% & 78.4\% & 71.9 \\
        NanoVerse-YOLO & \checkmark & \checkmark & \checkmark & 91.8\% & 88.9\% & 78.4\% & 71.9 \\
        \bottomrule
    \end{tabular}
    \label{tab:ablation_study}
\end{table*}
In the initial phase of the experiment, only NanoVerse-YOLO was used as a single-stage detector, with serial encoding to distinguish traffic sign categories. At this point, the model's performance on core metrics such as precision, recall, and mean average precision (mAP) was relatively weak. When a classifier trained with fine-grained text labels was introduced, all metrics showed significant improvement, initially confirming the positive impact of integrating textual semantics with visual features on model performance. Further replacing the text encoder with Rule-BERT resulted in another leap in model performance. This improvement is attributed to Rule-BERT's exceptional text parsing and encoding capabilities, which allow for more efficient extraction of semantic features, achieving deep coupling modeling with image features and significantly enhancing the accuracy and robustness of cross-modal representation. Finally, after integrating the semantic cache, the frames per second (FPS) processed by the classifier increased from 5 to 72. This result indicates that the semantic cache intelligently omits redundant feature encoding, significantly reducing computational overhead while achieving a substantial improvement in inference efficiency without sacrificing recognition accuracy.

\section{Conclusion}
This study addresses the core challenges of long-tail data distribution and small object detection in autonomous driving traffic sign recognition by proposing a two-stage algorithmic framework that integrates open vocabulary detection and cross-modal contrastive learning. In the detection phase, we innovatively design the NanoVerse-YOLO algorithm, which enhances the ability to capture features of small targets by integrating a re-parameterizable Visual-Language Path Aggregation Network (RepVL-PAN) with the SPD-Conv module, while optimizing localization accuracy through the use of Inner-WIoU loss. 

In the classification phase, we construct a cross-modal contrastive learning model, TSR-MCL, which leverages Vision Transformer and Rule-BERT to extract dual-modal features. This model is supported by a self-constructed cross-modal paired dataset, TSTIAD, consisting of 24,715 images covering 221 categories, effectively addressing the class confusion issues arising from long-tail distributions.

Experimental validation based on the TT100K dataset demonstrates that the model achieves a precision of 91.8\% and a recall of 88.9\%, with mAP50 and mAP50:95 reaching 78.4\% and 71.9\%, respectively, significantly surpassing current mainstream algorithms. In long-tail category detection, for rare categories with fewer than 10 instances, over 90\% of the categories achieve an mAP@0.5 exceeding 70\%, showing substantial improvements over existing long-tail detection methods. Additionally, by introducing a semantic cache, the inference speed of the classification module is enhanced from 5 FPS to 72 FPS, achieving a balance between accuracy and real-time performance. Future work will focus on model lightweighting, adaptability to adverse environments, and optimization of semantic reasoning capabilities to advance the practical application of traffic sign perception technology in autonomous driving scenarios.

\bibliographystyle{IEEEtran}
\bibliography{ref}
\vspace{-1cm}
\begin{IEEEbiography}[{\includegraphics[width=1in,height=1.25in,clip,keepaspectratio]{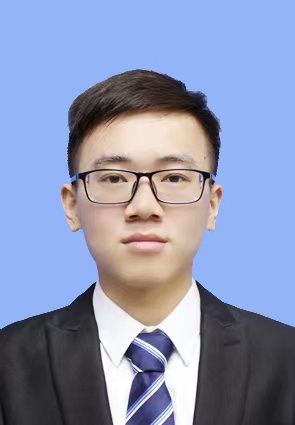}}]{Qiang Lu} received the M.S. degree in Agricultural Mechanization from South China Agricultural University in 2018. He is currently pursuing his Ph.D. degree in Computer Science and Technology (Intelligent Transportation Engineering) at the School of Intelligent Systems Engineering, Sun Yat-sen University. His research interests include Traffic Anomaly Detection, Traffic Behavior Understanding, and Intelligent Transportation Systems. 
\end{IEEEbiography}
\vspace{-1cm}
\begin{IEEEbiography}[{\includegraphics[width=1in,height=1.25in,clip,keepaspectratio]{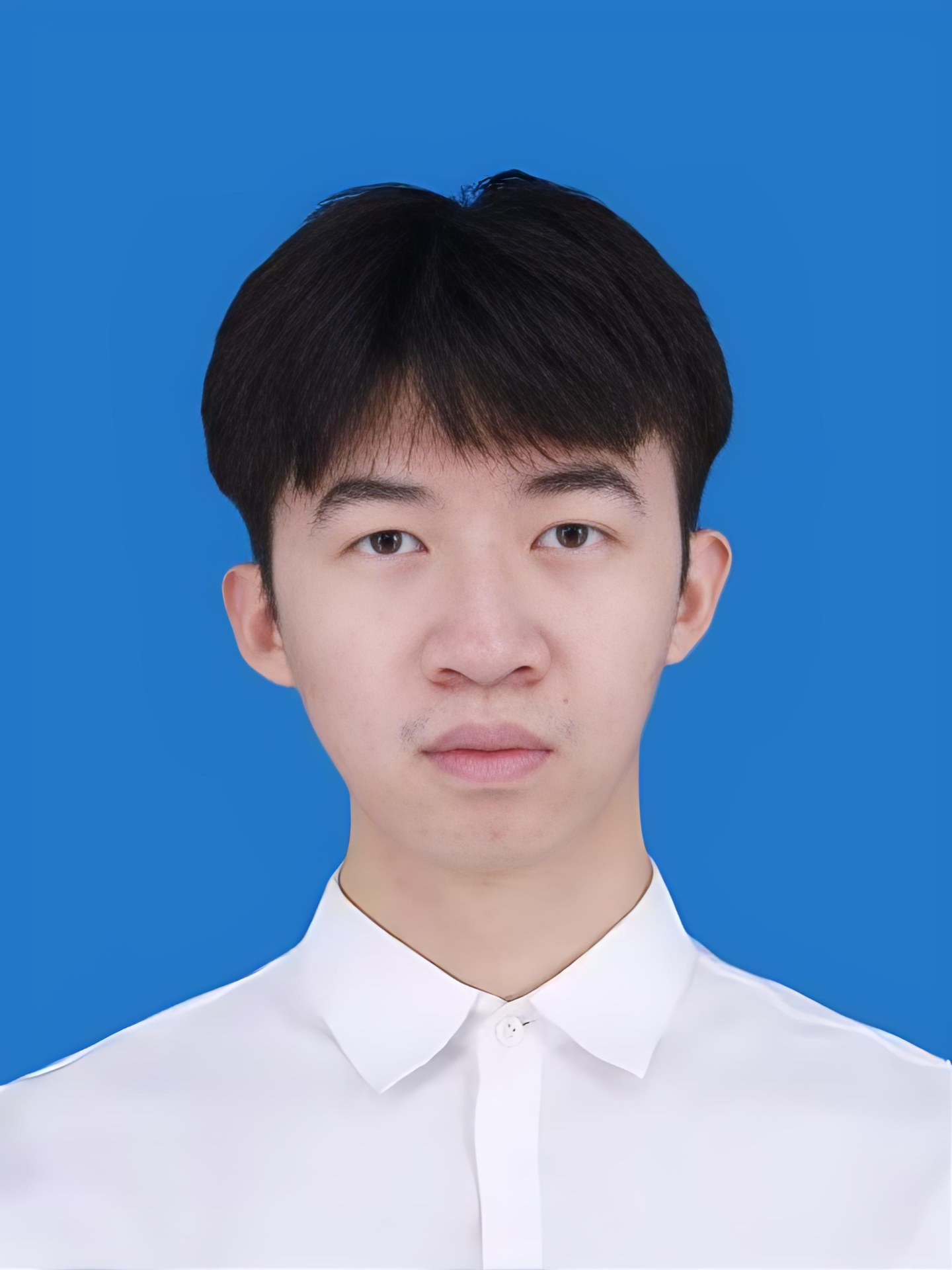}}]{Waikit Xiu} graduated with a Bachelor of Engineering degree in Transportation Engineering from Sun Yat-sen University, China, in 2025. He is currently pursuing a Master of Science degree in Artificial Intelligence at the University of Hong Kong. His research interests include multi-modal fusion, embodied intelligence, and intelligent transportation systems.
\end{IEEEbiography}
\vspace{-1cm}
\begin{IEEEbiography}[{\includegraphics[width=1in,height=1.25in,clip,keepaspectratio]{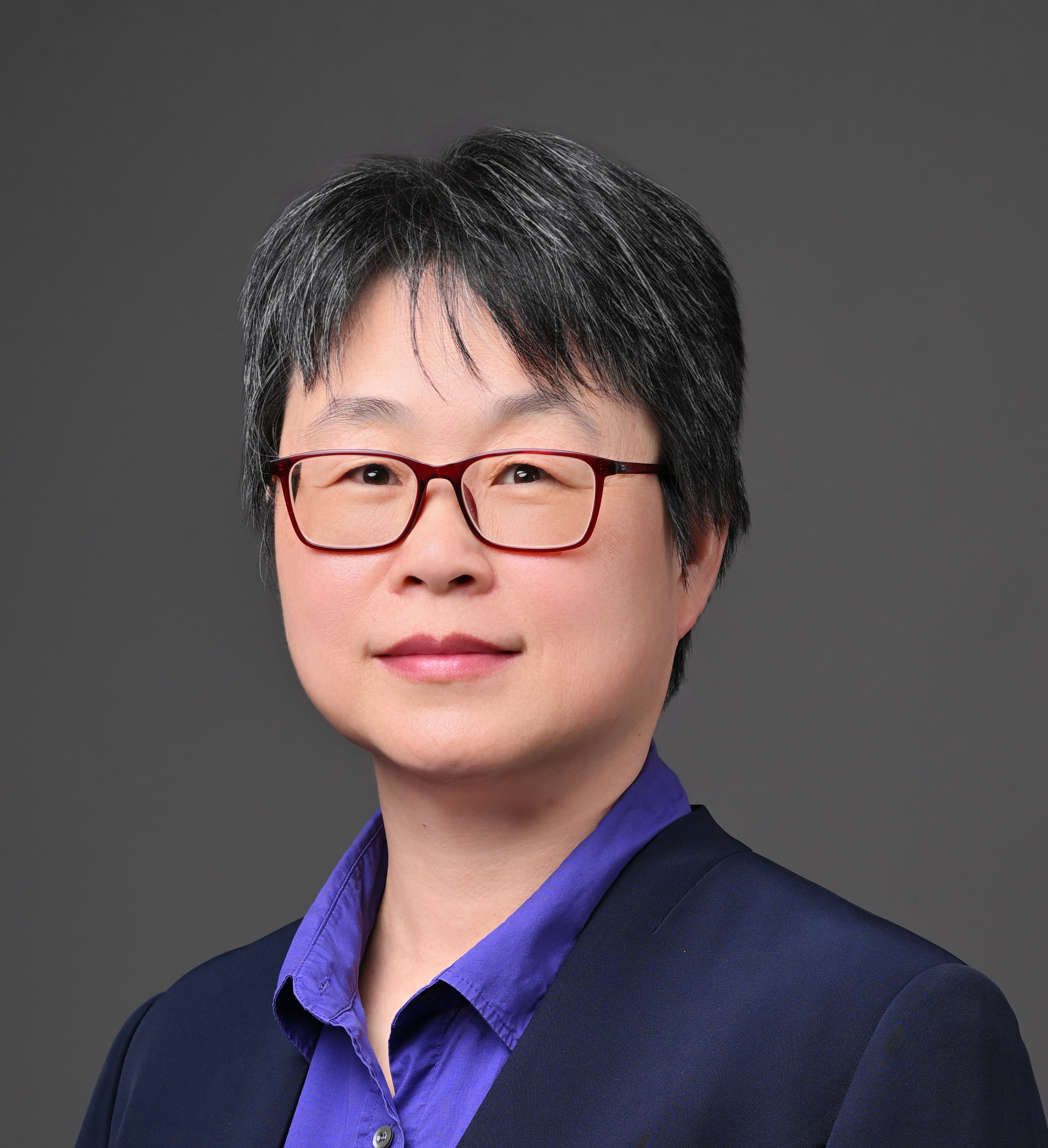}}]{Xiying Li} received the Ph.D. degree in optical engineering from Beijing Institute of Technology in 2002. She is currently a Professor with the School of Intelligent Systems Engineering, Sun Yat-sen University. Her research interests include intelligent transportation systems, traffic information collection, traffic video, and image big data processing and application.
\end{IEEEbiography}
\vspace{-1cm}
\begin{IEEEbiography}[{\includegraphics[width=1in,height=1.25in,clip,keepaspectratio]{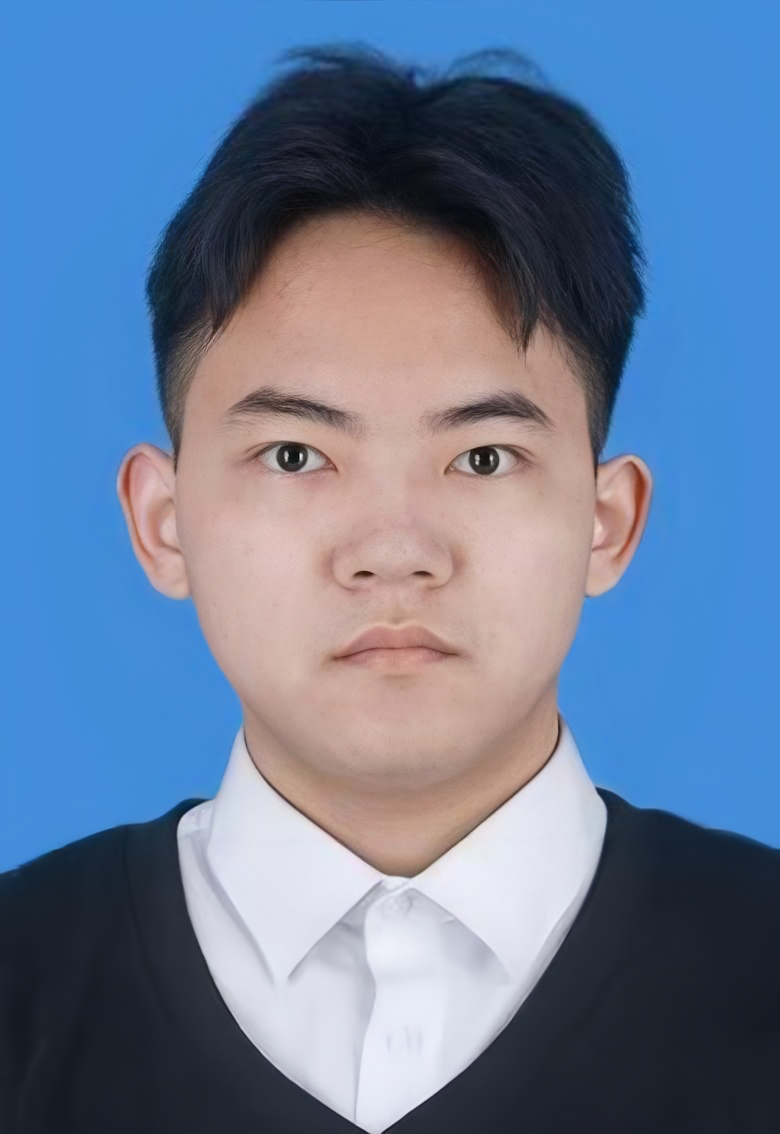}}]{Shenyu Hu} received the Bachelor of Engineering degree in Intelligent Science and Technology from Sun Yat-sen University, China, in 2025. His research interests include Video Image Technology Detection and Embodied Intelligence.
\end{IEEEbiography}
\vspace{-1cm}
\begin{IEEEbiography}[{\includegraphics[width=1in,height=1.25in,clip,keepaspectratio]{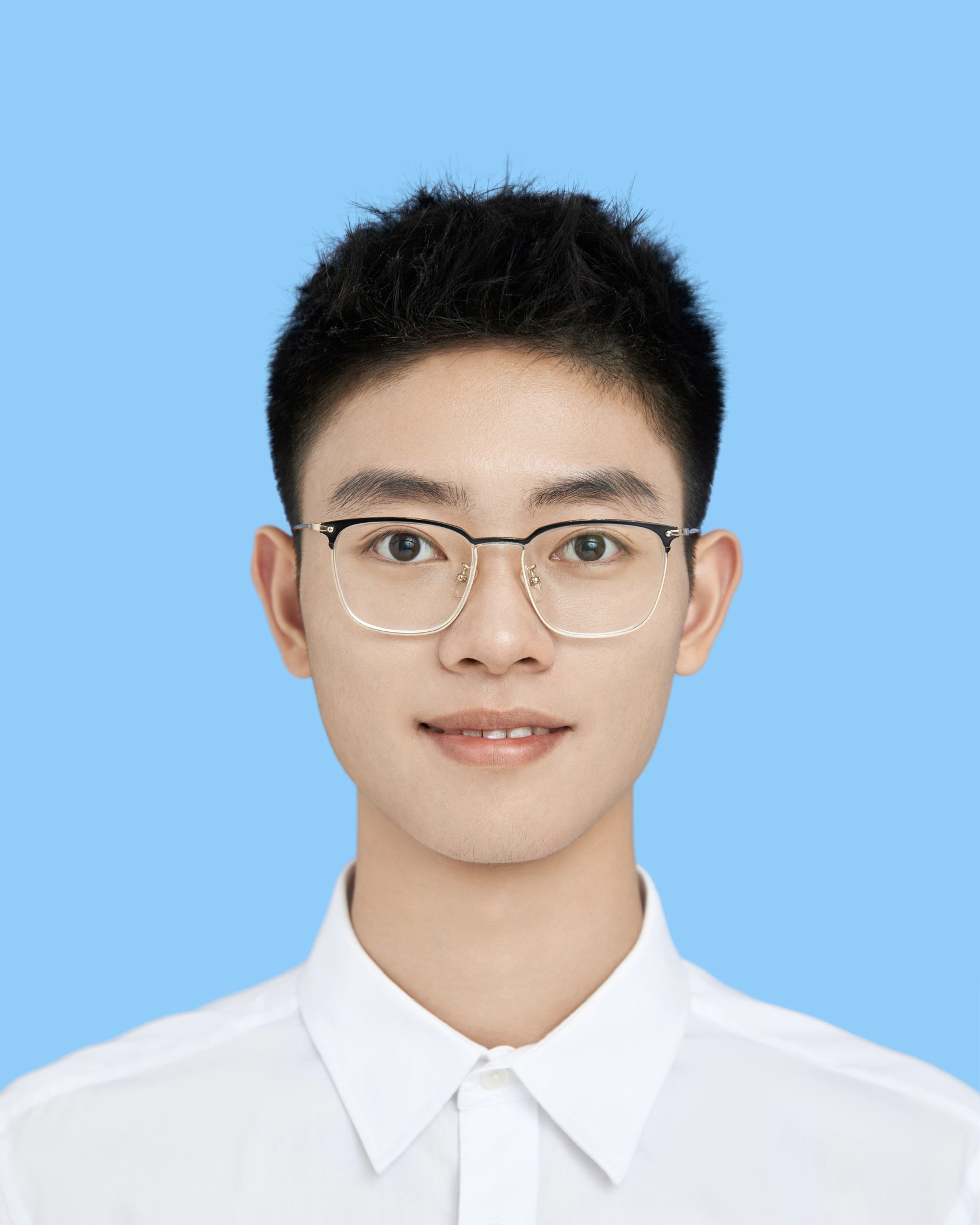}}]{Shengbo Sun} is currently pursuing the Bachelor’s degree in transportation engineering at Sun Yat-sen University, China. His research interests include traffic image processing, deep learning and computer vision.
\end{IEEEbiography}

\end{document}